\documentclass{article}
\usepackage{iclr2026_conference,times}

\usepackage{amsmath,amsfonts,bm}

\def\eqref#1{equation~\ref{#1}}

\def\1{\bm{1}}

\def\rva{{\mathbf{a}}}

\def\rvc{{\mathbf{c}}}

\def\rve{{\mathbf{e}}}

\def\rvh{{\mathbf{h}}}

\def\rvm{{\mathbf{m}}}

\def\rvo{{\mathbf{o}}}

\def\rvx{{\mathbf{x}}}

\DeclareMathAlphabet{\mathsfit}{\encodingdefault}{\sfdefault}{m}{sl}
\SetMathAlphabet{\mathsfit}{bold}{\encodingdefault}{\sfdefault}{bx}{n}

\def\gD{{\mathcal{D}}}

\def\gN{{\mathcal{N}}}

\usepackage{hyperref}
\usepackage{url}
\usepackage{duckuments}
\usepackage{adjustbox}
\usepackage{array}
\usepackage{booktabs}
\usepackage{subcaption}
\usepackage{lipsum}
\usepackage{amsmath}
\usepackage{float}
\usepackage{multirow}
\usepackage{enumitem}
\usepackage{listings}
\usepackage{algorithm}
\usepackage{algpseudocode}
\usepackage{bibunits}
\usepackage{changepage}
\usepackage{cleveref}
\usepackage{makecell}
\usepackage[table]{xcolor}

\definecolor{cornellred}{rgb}{0.7, 0.11, 0.11}
\definecolor{cadmiumgreen}{rgb}{0.0, 0.42, 0.24}
\definecolor{aliceblue}{rgb}{0.91, 0.94, 0.97}
\definecolor{darkblue}{rgb}{0.83, 0.89, 0.97}
\definecolor{Red7}{rgb}{0.941, 0.243, 0.243}
\definecolor{Green7}{RGB}{55, 178, 77}
\definecolor{Blue9}{rgb}{0.098,0.3,0.9}

\usepackage{pifont}
\newcommand{\cmark}{\ding{51}}%
\newcommand{\xmark}{\ding{55}}%
\newcommand{\ck}{\color{Green7}{\cmark}}
\newcommand{\xk}{\color{Red7}{\xmark}}

\definecolor{baselinecolor}{HTML}{EEEEEE}

\hypersetup{
  linkcolor = cornellred,
  citecolor  = cadmiumgreen,
  colorlinks = true,
  urlcolor = Blue9
}

\title{ContextVLA: Vision-Language-Action Model \\ with Amortized Multi-Frame Context}

\author{Huiwon Jang$^{1,2}${\quad}Sihyun Yu$^{1}${\quad}Heeseung Kwon$^{1,2}${\quad}Hojin Jeon$^{1}$\\
\textbf{Younggyo Seo$^{3}$\thanks{Equal advising. \hfill \textbf{Project page:} {\scriptsize\url{https://huiwon-jang.github.io/contextvla}}}{\quad}Jinwoo Shin$^{1,2\ast}$}\\
$^1$KAIST\,\,\,$^2$RLWRLD\,\,\,$^3$UC Berkeley
}

\newcommand{\sname}{ContextVLA}

\iclrfinalcopy

\begin{document}

\maketitle

\vspace{-0.2in}

\begin{abstract}
Leveraging temporal context is crucial for success in partially observable robotic tasks. However, prior work in behavior cloning has demonstrated inconsistent performance gains when using multi-frame observations. In this paper, we introduce \sname, a policy model that robustly improves robotic task performance by effectively leveraging multi-frame observations. Our approach is motivated by the key observation that Vision-Language-Action models (VLA), \textit{i.e.}, policy models built upon a Vision-Language Model (VLM), more effectively utilize multi-frame observations for action generation. This suggests that VLMs' inherent temporal understanding capability enables them to extract more meaningful context from multi-frame observations. However, the high dimensionality of video inputs introduces significant computational overhead, making VLA training and inference inefficient. To address this, \sname{} compresses past observations into a single context token, allowing the policy to efficiently leverage temporal context for action generation. Our experiments show that \sname~consistently improves over single-frame VLAs and achieves the benefits of full multi-frame training but with reduced training and inference times.

\end{abstract}

\begin{figure}[h]
\vspace{-0.31in}
\centering
\begin{subfigure}[t]{0.41\textwidth}
  \centering
  \includegraphics[width=.9\linewidth,height=0.348\linewidth]{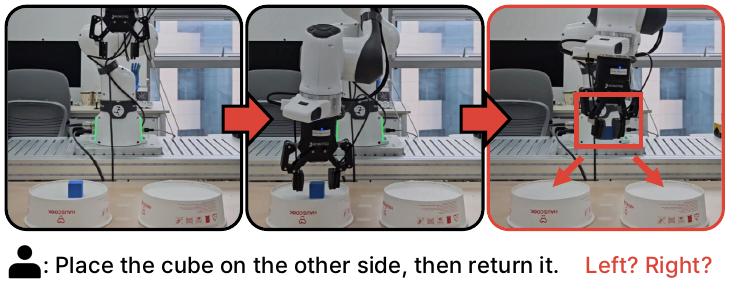}
\vspace{-0.1in}
  \subcaption{A task that requires temporal context}
  \label{fig:task_temporal_context}
\end{subfigure}
~
\begin{subfigure}[t]{0.28\textwidth}
  \centering
  \includegraphics[width=.828\linewidth]{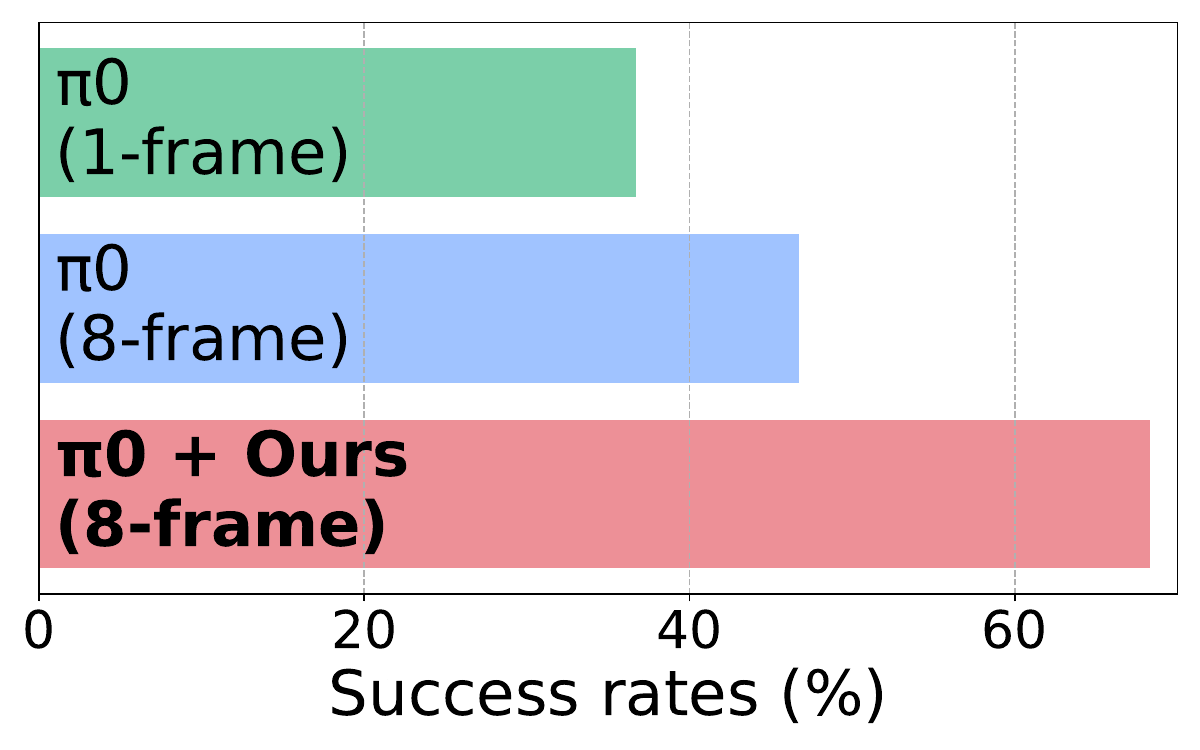}
\vspace{-0.1in}
  \subcaption{Real-world task result}
  \label{fig:xx}
\end{subfigure}
\begin{subfigure}[t]{0.28\textwidth}
  \centering
  \includegraphics[width=0.828\linewidth]{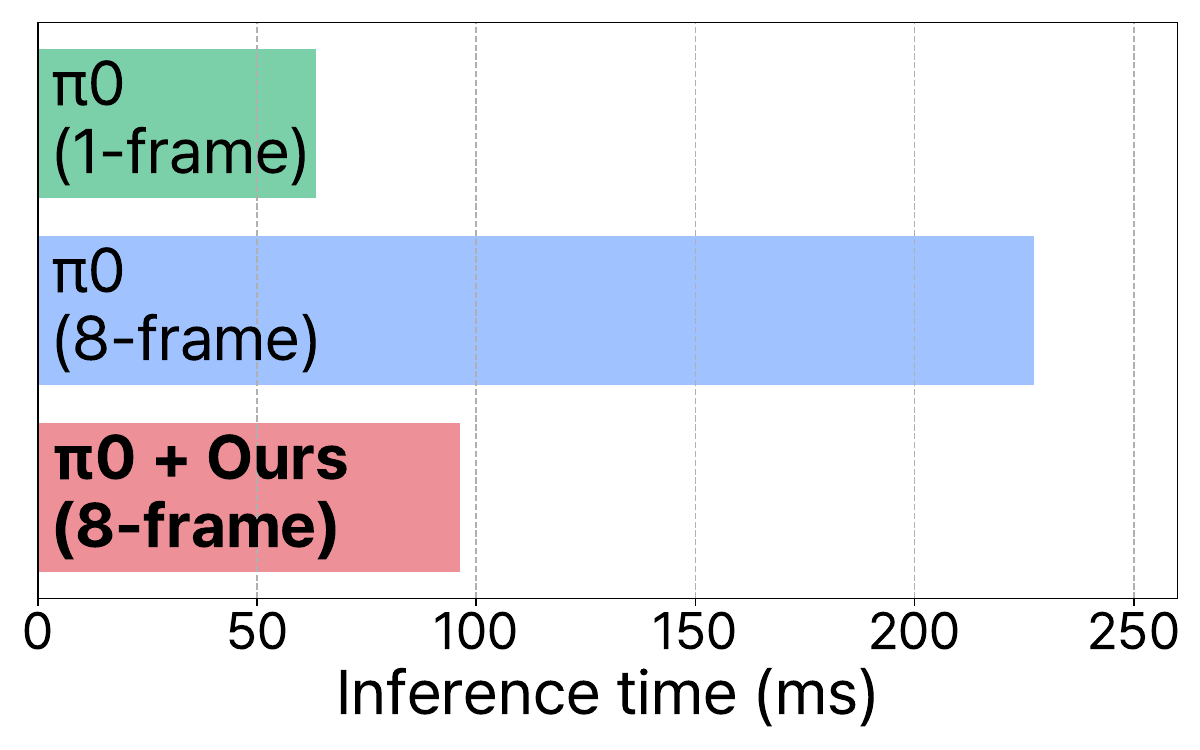}
\vspace{-0.1in}
  \subcaption{Inference efficiency}
  \label{fig:yy}
\end{subfigure}
\vspace{-0.15in}
\caption{\textbf{Overview.} (a) Many robotic tasks require temporal context to generate accurate actions. (b) By leveraging multi-frame observations, our proposed method, \sname, achieves higher averaged success rates (\%) over all baseline policies on real-world robotic tasks. (c) Moreover, our framework gets benefits of multi-frame training with reduced inference latency.
}
\label{fig:paper_overview}
\vspace{-0.06in}
\end{figure}

\vspace{-0.1in}

\section{Introduction}
\label{sec:intro}

Many robotic tasks are inherently non-Markovian, \textit{i.e.}, the optimal decision at a given timestep $t$ cannot be determined from the latest observation $\rvo_t$ alone but requires past sequential observations $\rvo_{1:t}$ \citep{kaelbling1998planning,zheng2024tracevla,shi2025memoryvla}. For instance, an object may become occluded during manipulation \citep{shi2025memoryvla}. Solving long-horizon tasks may also require context about the previous motions of a robot, and handling dynamic environments often involves tracking the motion trajectories of moving objects \citep{zhang2024uni,nasiriany2024robocasa}. Consequently, policy models must have capability to predict the actions based on the understanding of consecutive input observations (\textit{i.e.}, multi-frame observations) to perform real-world challenging task.

Despite its importance, recent behavior cloning (BC) policies usually have been trained with only a single frame observation \citep{kim2024openvla,bjorck2025gr00t,pertsch2025fast,nvidia2025gr00t}. This is mainly due to the mixed results reported in recent studies on training policy models with multi-frame observations. Specifically, several works argue that multi-frame observations do improve performance \citep{wu2023unleashing,team2024octo,cheang2024gr,zheng2024tracevla,liu2025towards}, but surprisingly, many others have observed contradictory results; namely, this training scheme can even lead to performance degradation \citep{de2019causal,wen2020fighting,spencer2021feedback,seo2023regularized,torne2025learning}.

\begin{figure}[t]
\centering
\begin{subfigure}[t]{0.39\textwidth}
  \centering
  \includegraphics[width=0.90\linewidth]{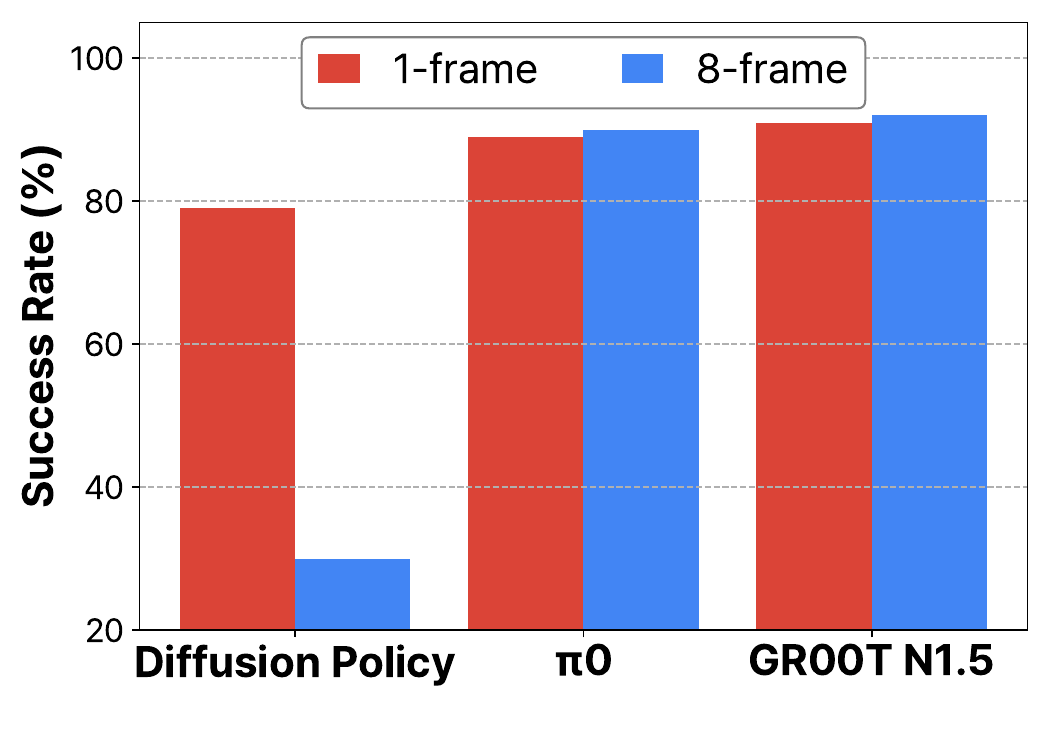}
  \subcaption{Performance of recent policy models trained using either 1-frame or 8-frame observations.}
  \label{fig:init_robomimic}
\end{subfigure}\hfill
\begin{subfigure}[t]{0.56\textwidth}
  \centering
  \includegraphics[width=0.90\linewidth]{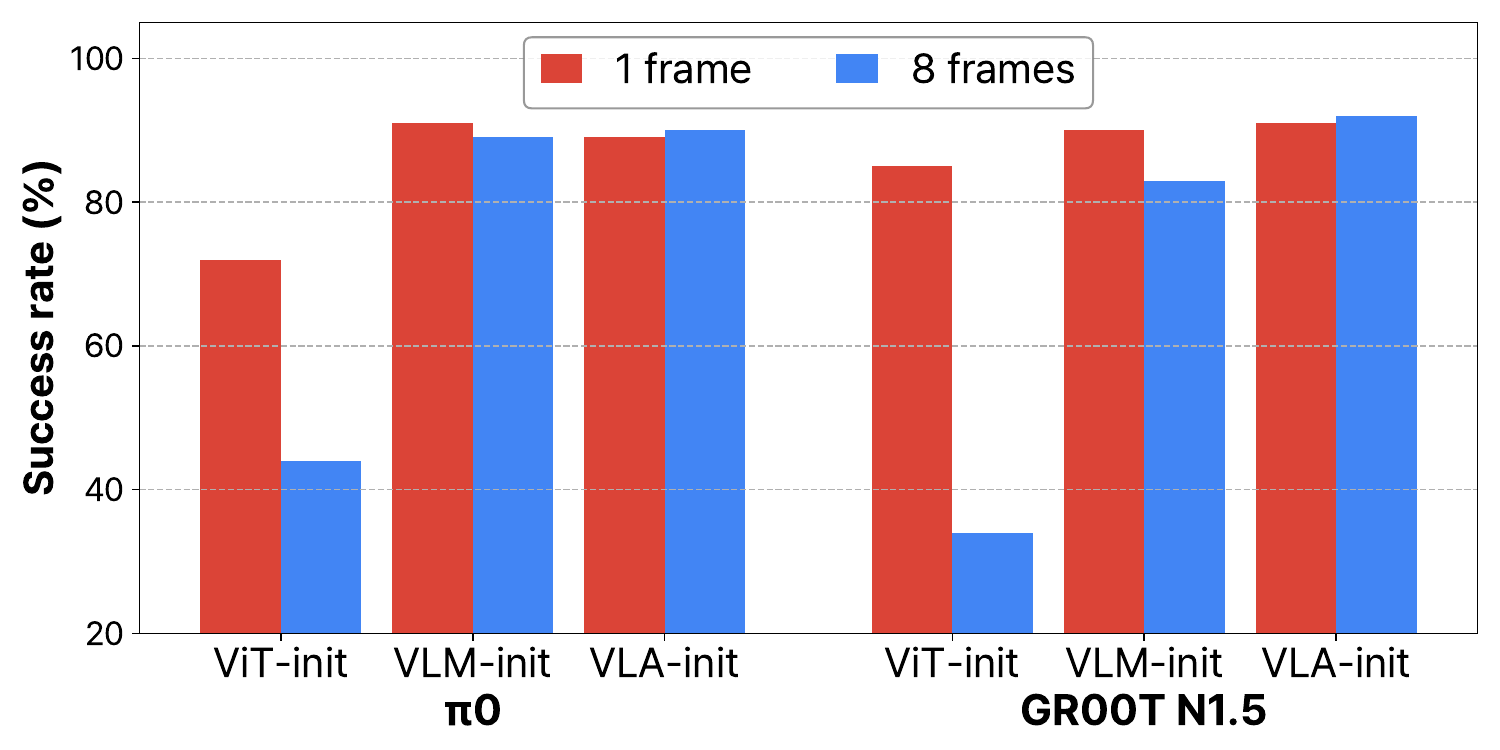}
  \subcaption{Performance of policy models of VLA architecture trained using either 1-frame or 8-frame observations under different weight initialization schemes.}
  \label{fig:init_libero}
\end{subfigure}
\caption{\textbf{Effect of multi-frame observations for training various policy models.} We report the success rates (\%) of various policy models fine-tuned on Square task from the Robomimic benchmark \citep{mandlekar2021matters}. (a) When training policy models using multi-frame observations, traditional policy model (Diffusion policy) shows significant performance degradation, whereas recent Vision-Language-Action models (VLA; $\pi_0$ and GR00T N1.5) do not. (b) We find that the key factor in overcoming this problem is leveraging a pre-trained Vision-Language Model (VLM) to extract temporal information for action generation. ViT, VLM, and VLA-init indicate how the VLA architecture is initialized for training; we use a pre-trained vision encoder, VLM, or VLA, respectively, and other parameters are randomly initialized.
}
\vspace{-0.1in}
\label{fig:motivation}
\end{figure}
\vspace{-0.1in}
\paragraph{Contribution.} We introduce a framework that enables BC policy models to effectively leverage multi-frame observations, thereby achieving consistent performance improvement across a wide range of manipulation tasks. To this end, we start by performing an analysis of why prior works have reported inconsistent gains from multi-frame observations. We find that while standard policies often suffer from performance degradation with multi-frame data, Vision-Language-Action models (VLA; \citealt{black2024pi_0,bjorck2025gr00t}), \textit{i.e.}, policy models initialized from or conditioned on a Vision-Language Model (VLM; \citealt{beyer2024paligemma, chen2025eagle}), mitigate this problem (\Cref{fig:init_robomimic}). In particular, our analysis shows that the VLM serves as the key component in mitigating performance degradation (\Cref{fig:init_libero}). This finding suggests that the VLM's temporal understanding is key to extracting more meaningful context from videos for action generation. However, leveraging multi-frame observations for VLA training and inference significantly increases computational cost, as high-dimensional sequences must be processed by large VLMs (often $>$1B parameters). Thus, it is important to develop efficient approaches for exploiting multi-frame information with VLAs.

Based on this analysis, we propose \sname, an efficient framework that leverages a VLM's temporal understanding to learn a BC policy model that utilizes multi-frame observations. Our key idea is to compress past observations into a single context token, which enables the VLA to efficiently capture temporal context (\textit{e.g.}, the movement of a robot) to generate actions with reduced computation overhead. Specifically, \sname{} first processes the full observation sequence using the initial blocks of the VLM backbone. It then aggregates the tokens from past observations into a single context token. Then the remaining VLM blocks process the sequence consisting of this new context token and the tokens for the current observation. After that, the resulting VLM features are fed into an action decoder to generate actions in either an autoregressive \citep{pertsch2025fast} or diffusion-based manner \citep{black2024pi_0,bjorck2025gr00t}.

We verify the effectiveness of our scheme through extensive experiments on various robotic manipulation benchmarks, including Libero \citep{liu2023libero}, Simpler-WidowX \citep{li2024evaluating}, and Robocasa \citep{nasiriany2024robocasa}. Our results show that \sname~consistently improves the performance of recent state-of-the-art VLAs that use single-frame observations. For instance, on the Simpler-WidowX benchmark, \sname~improves the average success rate of $\pi_0$ \citep{black2024pi_0} by 14.4\%  (41.8\% → 56.2\%). Moreover, we find that \sname{}~is particularly effective on long-horizon real-world robotic tasks that require temporal understanding (Figure~\ref{fig:env_nonmark}). For example, fine-tuning $\pi_0$ \citep{black2024pi_0} with our framework achieves a 65\% success rate on the pick-and-place twice (PnP Twice) task, whereas the single-frame baseline gets only 25\%. Finally, we show that \sname~enables efficient training and inference of multi-frame VLA models, delivering the benefits of multi-frame training while significantly reducing the training and inference costs.

\begin{figure}[t]
\centering
\includegraphics[width=0.9\textwidth]{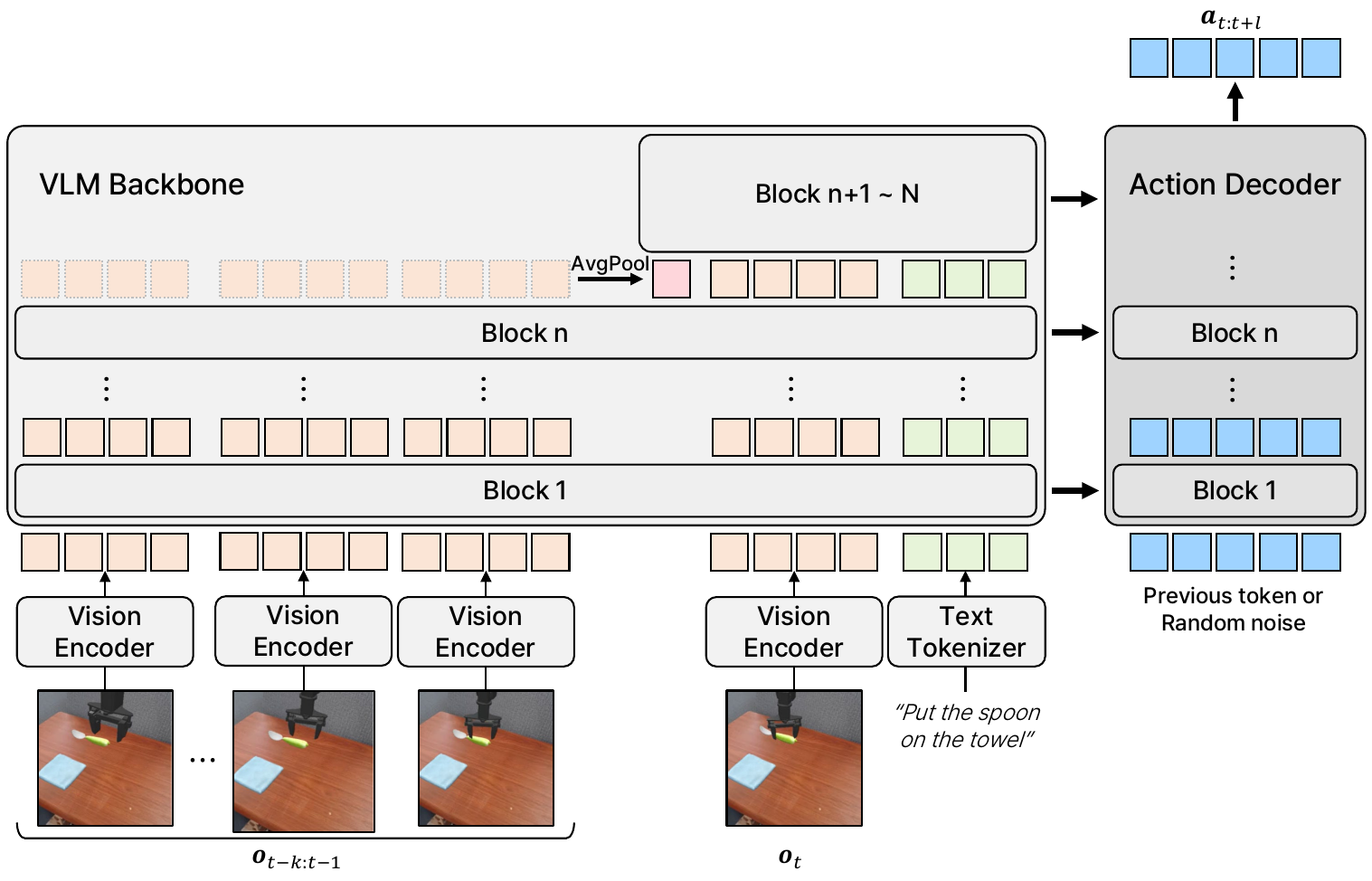}
\vspace{-0.1in}
\caption{\textbf{Overview of \sname.} We design an efficient Vision-Language-Action model (VLA) that generates actions using multi-frame visual observations. We use a Vision-Language Model (VLM) to encode observations $\rvo_{t-k:t}$, where we compress past observations $\rvo_{t-k:t-1}$ into a single context token $\rvm$ at the VLM block $n$. We then leverage the VLM features to generate actions via either autoregressive modeling or diffusion-based modeling.}
\label{fig:overview}
\vspace{-0.1in}
\end{figure}
\section{Method}
\label{sec:method}
In this section, we introduce \sname, an efficient training framework for Vision-Language-Action models (VLA) that leverages \emph{multi-frame} observations. In a nutshell, \sname~encodes hidden states of past visual observations into a compact global context token at the intermediate layer of Vision-Language-Model (VLM) backbone, while preserving the number of tokens of current observations. After that, these VLM features are fed into an action decoder to generate the action. We provide the overview of \sname~in \Cref{fig:overview}.

\subsection{Preliminaries}

\paragraph{Problem Setup.}
Let $\tau=\{(\rvo_t, \rvc_t, \rva_t)\}_{t=1}^T$ be an expert demonstration consisting of visual observation $\rvo_t$, language instruction $\rvc_t$, and robot action $\rva_t$ for each timestep $t$, and let $\gD = \{\tau_i\}_{i=1}^{N}$ be a dataset consisting of expert demonstrations. Here, visual observation $\rvo_t$ can be either single or multi-view depending on the environment. Moreover, we denote by $\rvx_{a:b}$ the sequence of consecutive vectors $[\rvx_a, \ldots, \rvx_b]$ for $a < b$.

Given $\tau \in \gD$, our goal is to train a policy model $\pi_\theta(\rva_{t:t+l}| \rvo_{t-k:t},\rvc_t)$, which predicts $l+1$ future actions $\rva_{t:t+l}$ of the robot  \citep{zhao2023learning,chi2023diffusion} from $k+1$-frame observations $\rvo_{t-k:t}$ and a language instruction $\rvc_t$. In particular, we aim this policy $\pi_\theta$ to leverage $k+1$-frame visual observations so that it understands not only the current state, but the context of past observations to generate actions. 
However, using longer past observations can dramatically increase computation and memory overheads by increasing input dimensionality that $\pi_\theta$ processes, resulting in inefficient training and inference. Therefore, we design $\pi_\theta$ to process past observations efficiently.

\paragraph{Multi-frame Vision-Language-Action Model.}
We design the policy $\pi_\theta$ as VLA, where it encodes visual observations $\rvo_{t-k:t}$ and a task instruction $\rvc_t$ using a pre-trained VLM \citep{beyer2024paligemma,bai2025qwen2,chen2025eagle}, and uses the extracted features from the VLM to generate a robot action $\rva_{t:t+l}$. Action generation can be performed in either an autoregressive \citep{kim2024openvla,pertsch2025fast} or a diffusion-based manner \citep{black2024pi_0,bjorck2025gr00t}, conditioning on the VLM features. We describe the detailed action generation process in \Cref{appendix:action_generation}.

\subsection{\sname} \label{sec:method:vlm}

To leverage past observations for action generation, \sname~uses Vision-Language Model (VLM) to process multi-frame observations. However, video inputs contain many tokens, substantially increasing compute and memory overhead in the VLM. To address this, our key idea is to compress past observations into a single context token, allowing the VLM to capture the temporal context of past observations, \textit{e.g.}, movement of the robot, while reducing computation and memory overhead. Specifically, we compress the past observations at the intermediate layer of the VLM backbone by applying average pooling. And then, an action decoder generates robot actions conditioned on the resulting VLM features.

Formally, given multi-frame observations $\rvo_{t-k:t}$, we first process them all through a vision encoder $f$ to obtain visual features $\rve_{t-k:t} = f(\rvo_{t-k:t})$. We then use  $\rvx = [\rve_{t-k:t}, \rvc_t]$ \textit{i.e.}, a concatenation of $\rve_{t-k:t}$ and $\rvc_t$, as input tokens to the VLM backbone $g$.

\paragraph{Amortization of Past Observation.}
The next step is to process the input tokens $\rvx$ with the VLM backbone $g$ to extract the features that will be used to generate the action. To efficiently process multi-frame observations using the VLM backbone, we compress past observations into a single context token within the intermediate layer of the VLM model $g$.

Formally, let $N$ be the number of VLM backbone blocks. We split the VLM backbone into two parts at the $n$-th block. First, in the first $n$ blocks, we process all tokens of $\rvx$ through the VLM blocks to get intermediate hidden states $\rvh = [\rvh_{t-k:t}, \rvh_\rvc]$. In particular, when tokens are fed into the self-attention layers, we apply a causal mask to the visual tokens, regardless of the original VLM attention mask. This allows efficient inference by processing and caching past observations before the next timestep. After this, we compress the past visual observations into a context token using average pooling, \textit{i.e.}, $\rvm = \texttt{AvgPool}([\rvh_{t-k:t-1}])$ to capture temporal context of past observations. In the remaining $N-n$ blocks, we replace the hidden states of the past observations $\rvh_{t-k:t-1}$ with the context token $\rvm$, and process $[\rvm, \rvh_t, \rvh_\rvc]$ through the blocks.
As a result, we obtain the VLM features, which encode both the current observation and an amortized context of past observations.

\paragraph{Action Decoder.}
Our action decoder generates a chunk of robot action $\rva_{t:t+l}$ with a length $l+1$ \citep{zhao2023learning,chi2023diffusion}, using VLM features as conditioning. Because our amortization scheme does not depend on the type of action decoders, \sname~can be applied to any VLAs regardless of their action decoder models. Specifically, it can be applied to autoregressive \citep{kim2024openvla,bu2025univla,pertsch2025fast} or diffusion-based action decoders \citep{black2024pi_0,bjorck2025gr00t,nvidia2025gr00t}. For instance, for the autoregressive modeling, we encode an action $\rva_{t:t+l}$ into discrete action tokens using the action tokenizer \citep{bu2025univla,pertsch2025fast}, and then use the same VLM backbone to generate the discrete action tokens via next action token prediction. For the diffusion-based modeling, we generate an action using a diffusion transformer (DiT \citet{peebles2022scalable}) conditioning on the VLM hidden states, \textit{e.g.}, output tokens of the VLM or key-values in the VLM blocks.

\paragraph{Training Objective.} We train \sname~to predict the target ground-truth action $\rva_{t:t+l}$ in the expert trajectory $\tau =\{(\rvo_t, \rvc_t, \rva_t)\}_{t=1}^T$. Specifically, we train the model $\pi_\theta$ to minimize the loss 
$\ell (\pi_\theta(\rvo_{t-k:t}, \rvc_t), \rva_{t:t+l})$, where $\ell$ corresponds to either a next-token prediction loss for the autoregressive action modeling or a flow-matching loss for the diffusion-based action modeling.

\paragraph{Efficient Inference via KV-caching.}
\sname~generates actions faster than a VLA that uses videos without compression, since we use an amortized token instead of past observations in most VLM blocks. In addition to this, 
we further make an inference of \sname~faster by processing as many of the observations as possible before the next timestep.
Specifically, at timestep $t-1$, we process $\rvo_{t-k:t-1}$ through the first $n$ VLM blocks to obtain a KV-cache for each block, and obtain a context token $\rvm$. At timestep $t$, since we explicitly implement the attention layers of the first $n$ VLM blocks with causal-attention, we generate actions using $\rvo_t$, pre-computed $\rvm$, and the KV-cache, rather than re-processing $\rvo_{t-k:t-1}$.
\begin{figure}[t]
\centering
\begin{subfigure}[t]{0.48\textwidth}
  \centering
  \includegraphics[width=\linewidth]{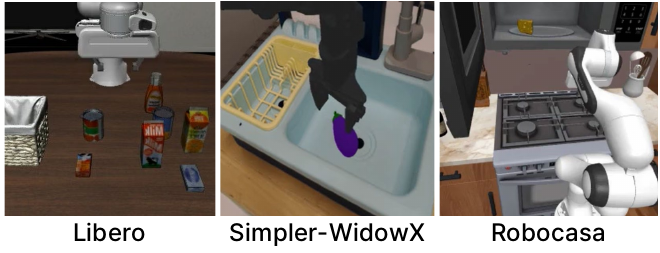}
  \subcaption{Simulated robotic manipulation tasks}
  \label{fig:env_std}
\end{subfigure}
\begin{subfigure}[t]{0.48\textwidth}
  \centering
  \includegraphics[width=\linewidth]{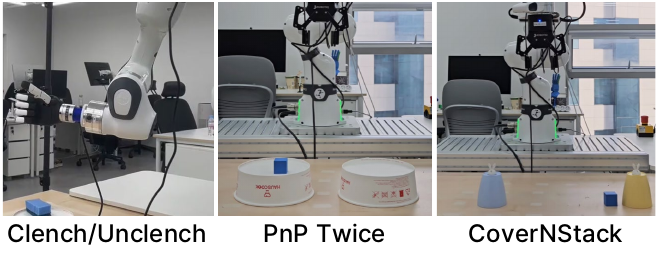}
  \subcaption{Real-world robotic tasks}
  \label{fig:env_nonmark}
\end{subfigure}
\caption{\textbf{Examples of visual observations from the evaluation tasks.} (a) We consider simulated robotic manipulation tasks from Libero \citep{liu2023libero}, Simpler-WidowX \citep{li2024evaluating}, and Robocasa \citep{nasiriany2024robocasa}. (b) We design real-world robotic tasks: Clench/unclench hand (Clench/Unclench), pick-and-place twice (PnP Twice), and cover and stack (CoverNStack).}
\label{fig:environment_examples}
\end{figure}

\section{Experiments}
\label{sec:exp}

We design experiments to investigate the following questions:
\begin{itemize}[leftmargin=*,itemsep=0mm]
    \item Can \sname~leverage the multi-frame observations to perform diverse robotic tasks without performance degradation? (\Cref{tab:main_libero,tab:main_simpler,tab:main_robocasa}) In particular, is \sname~effective on robotic tasks that particularly need past observations? (\Cref{tab:main_non_markovian})
    \item Is \sname~efficient during training and inference? (\Cref{tab:analysis_efficiency} and \Cref{fig:wall_time_clock})
    \item What is the effect of each component of \sname? (\Cref{tab:component_analysis}) 
\end{itemize}

\subsection{Experimental Setup}

\paragraph{Implementation Details.}
We report the performance of our method, \sname, by fine-tuning pre-trained Vision-Language-Action models (VLA).
Specifically, we use $\pi_0$ \citep{black2024pi_0}, GR00T N1.5 \citep{nvidia2025gr00t}, and $\pi_0$-FAST \citep{pertsch2025fast} by following their official implementation. For $\pi_0$ and $\pi_0$-FAST, we perform full-finetuning, \textit{i.e.}, we update all model parameters, but we freeze the vision encoder and VLM backbone for GR00T N1.5. We use 8 consecutive frames as multi-frame observations, and we compress past observations into context tokens at the output of the 2nd VLM block (\textit{i.e.}, $n$=2). We report the best performance by evaluating the models at fixed intervals during training. We provide more detailed implementation details in \Cref{appendix:implementation}.

\paragraph{Baselines.}
We compare the performance of \sname~with recent open-source VLAs: Octo \citep{team2024octo}, OpenVLA \citep{o2024open}, RoboVLMs \citep{liu2025towards}, TraceVLA \citep{zheng2024tracevla}, SpatialVLA \citep{qu2025spatialvla}, NORA \citep{hung2025nora}, $\pi_0$ \citep{black2024pi_0}, GR00T N1.5 \citep{nvidia2025gr00t}, and $\pi_0$-FAST \citep{pertsch2025fast}. In particular, to evaluate the benefit of \sname, we compare $\pi_0$, GR00T N1.5, and $\pi_0$-FAST trained and fine-tuned with single-frame inputs against their counterparts fine-tuned with multi-frame inputs using our method. All models are fine-tuned with the same batch size and number of iterations, and we report their best success rates at fixed evaluation intervals. We describe more details of baselines in \Cref{appendix:baselines}.

\subsection{Simulated Robotic Tasks}
To demonstrate our method can leverage multi-frame visual observations, we first evaluate our method by fine-tuning pre-trained VLAs on diverse simulated robotic manipulation benchmarks (see \Cref{fig:env_std} for examples of tasks from the benchmarks used in our experiments).

\paragraph{Experimental Setup.} 
We first consider Libero benchmark \citep{liu2023libero}, one of the popular benchmarks for evaluating VLA, which includes 4 different sub-benchmarks (Spatial, Object, Goal, and Long) that contain 10 tasks each. We also consider 4 tasks from the Simpler-WidowX benchmark \citep{li2024evaluating}, which is a more challenging setup due to a visual gap between real-world training data and simulated test environments (Real-to-Sim). We report the performance by fine-tuning the models on Bridge v2 dataset \citep{ebert2021bridge}. Moreover, we consider the Robocasa benchmark \citep{nasiriany2024robocasa}, which includes 24 tasks in simulated kitchen environments. It consists of more than 2500 different kitchen scenes across more than 150 object categories, requiring a policy to have instruction following and generalization ability over the scene and object. We report the performance by fine-tuning all models on the machine-generated training dataset, consisting of 100 demos per task. For each benchmark, we fine-tune all models for 60K iterations using the AdamW optimizer with a batch size of 32. We provide a more detailed experimental setup in \Cref{appendix:implementation} and an explanation about the benchmark with evaluation details in \Cref{appendix:task:simul}.

\begin{table*}[t]
\caption{\textbf{Results on Libero.} We report the success rates (\%) of various VLAs fine-tuned on the training dataset of Libero \citep{liu2023libero,kim2024openvla}. For $\pi_0$ \citep{black2024pi_0}, $\pi_0$-FAST \citep{pertsch2025fast}, and GR00T N1.5 \citep{nvidia2025gr00t}, we report the performance by fine-tuning the pre-trained model with the official implementations, 
and other reported numbers borrow from \citet{team2024octo,hung2025nora}.}
\centering\small
\vspace{-0.1in}
\begin{adjustbox}{max width=\textwidth}
    \begin{tabular}{lc ccccc}
        \toprule
        Method & \# frames & Spatial & Object & Goal & Long & Avg. \\
        \midrule
        Octo \citep{team2024octo} & 2 & 78.9 & 85.7 & 84.6 & 51.1 & 75.1 \\
        OpenVLA \citep{kim2024openvla} & 1 & 84.9 & 88.4 & 79.2 & 53.7 & 76.5 \\
        TraceVLA \citep{zheng2024tracevla} & 6 & 84.9 & 85.2 & 75.1 & 54.1 & 74.8 \\
        SpatialVLA \citep{qu2025spatialvla} & 1 & 88.2 & 89.9 & 78.6 & 55.5 & 78.1 \\
        NORA \citep{hung2025nora} & 1 & 92.2 & 95.4 & 89.4 & 74.6 & 87.9 \\
        \midrule
        $\pi_0$ \citep{black2024pi_0} & 1 & 96.0 & 97.2 & 96.0 & 89.2 & 94.6 \\
        \textbf{ + \sname~(Ours)} & 8 & \textbf{97.4} & \textbf{98.2} & \textbf{96.4} & \textbf{93.8} & \textbf{96.5} \\
        \midrule
        $\pi_0$-FAST \citep{pertsch2025fast} & 1 & 96.6 & 96.6 & 95.2 & 85.2 & 93.4 \\
        \textbf{ + \sname~(Ours)} & 8 & \textbf{98.3} & \textbf{99.2} & \textbf{95.6} & \textbf{90.2} & \textbf{95.8} \\
        \midrule
        GR00T N1.5 \citep{nvidia2025gr00t} & 1  & 98.3 & \textbf{99.4} & 96.7 & 89.0 & 95.9 \\
        \textbf{ + \sname~(Ours)} & 8  & \textbf{98.4} & 99.0 & \textbf{97.2} & \textbf{93.4} & \textbf{97.0} \\
        \bottomrule
    \end{tabular}\label{tab:main_libero} 
\end{adjustbox}
\end{table*}

\begin{table*}[t]
\caption{\textbf{Results on Simpler-WidowX.} We report the success rates (\%) of various VLAs fine-tuned on the Bridgev2 dataset \citep{walke2023bridgedata}. For $\pi_0$ \citep{black2024pi_0}, $\pi_0$-FAST \citep{pertsch2025fast}, and GR00T N1.5 \citep{nvidia2025gr00t}, we report the performance by fine-tuning the pre-trained model with the official implementations, 
and other reported numbers borrow from \citet{qu2025spatialvla}.}
\centering\small
\vspace{-0.1in}
\begin{adjustbox}{max width=\textwidth}
    \begin{tabular}{lc ccccc}
        \toprule
        Method & \# frames & \shortstack{Spoon\\on Towel} & \shortstack{Carrot\\on Plate} & \shortstack{Stack\\Cube} & \shortstack{Put Eggplant\\ in Basket} & Avg. \\
        \midrule
        Octo-base \citep{team2024octo} & 2 & 12.5 & \phantom{0}8.3 &  \phantom{0}0.0 & \phantom{0}43.1 & 16.0 \\
        Octo-small \citep{team2024octo} & 2 & 47.2 & \phantom{0}9.7 & \phantom{0}4.2 & \phantom{0}56.9 & 29.5 \\
        OpenVLA \citep{kim2024openvla} & 1 & \phantom{0}0.0 & \phantom{0}0.0 & \phantom{0}0.0 & \phantom{0}\phantom{0}4.1 & \phantom{0}1.0 \\
        RoboVLMs \citep{liu2025towards} & 16 & 29.2 & 25.0 & 12.5 & \phantom{0}58.3 & 31.3 \\
        SpatialVLA \citep{qu2025spatialvla} & 1 & 16.7 & 25.0 & 29.2 & 100.0 & 42.7 \\
        \midrule
        $\pi_0$ \citep{black2024pi_0} & 1 & 46.7 & 38.7 & \textbf{42.7} & \phantom{0}39.3 & 41.8 \\
        \textbf{ + \sname~(Ours)} & 8 & \textbf{53.3} & \textbf{56.0} & 41.3 & \phantom{0}\textbf{74.0} & \textbf{56.2} \\
        \midrule
        $\pi_0$-FAST \citep{pertsch2025fast} & 1 & 59.0 & 79.0 & 65.0 & \phantom{0}33.0 & 59.0 \\
        \textbf{ + \sname~(Ours)} & 8 & \textbf{60.7} & \textbf{81.3} & \textbf{78.7} & \phantom{0}\textbf{62.0} & \textbf{70.7} \\
        \midrule
        GR00T N1.5 \citep{nvidia2025gr00t} & 1 & \textbf{30.0} & 28.0 & \textbf{16.0} & \phantom{0}42.7 & 29.2 \\
        \textbf{ + \sname~(Ours)} & 8 & 28.0 & \textbf{29.3} & 14.7 & \phantom{0}\textbf{50.3} & \textbf{31.8} \\
        \bottomrule
    \end{tabular}\label{tab:main_simpler} 
\end{adjustbox}
\end{table*}

\begin{table*}[t]
\caption{\textbf{Results on Robocasa.} We report the success rates (\%) of various VLAs fine-tuned on the training dataset of Robocasa \citep{nasiriany2024robocasa}, consisting of 24 tasks with 100 demos per task. We report the performance by fine-tuning the pre-trained model with the official implementations.}
\centering\small
\vspace{-0.1in}
\begin{adjustbox}{max width=\textwidth}
    \begin{tabular}{lc ccc}
        \toprule
        Method & \# frames & Pick and Place & Others & Avg. \\
        \midrule
        $\pi_0$ \citep{black2024pi_0} & 1 & 33.3 & 68.8 & 57.0 \\
        \textbf{ + \sname~(Ours)} & 8 & \textbf{34.8} & \textbf{70.6} & \textbf{58.7} \\
        \midrule
        $\pi_0$-FAST \citep{pertsch2025fast} & 1 & 45.5 & 69.0 & 60.2 \\
        \textbf{ + \sname~(Ours)} & 8 & \textbf{48.5} & 69.0 & \textbf{62.2} \\
        \midrule
        GR00Tn1.5 \citep{nvidia2025gr00t} & 1 & 50.3 & 68.8 & 62.6 \\
        \textbf{ + \sname~(Ours)} & 8 & \textbf{53.0} & \textbf{69.9} & \textbf{64.3} \\
        \bottomrule
    \end{tabular}\label{tab:main_robocasa} 
\end{adjustbox}
\end{table*}

\begin{table*}[t]
\caption{\textbf{Results on real-world robotic tasks.} We report the success rates (\%) of various VLAs fine-tuned on the training dataset of each task. We report the performance by fine-tuning the pre-trained model with the official implementations.}
\centering\small
\vspace{-0.1in}
\begin{adjustbox}{max width=\textwidth}
    \begin{tabular}{lc ccccc}
        \toprule
        &  & & \multicolumn{2}{c}{PnP Twice} & \multicolumn{2}{c}{CoverNStack} \\
        \cmidrule(lr){4-5} \cmidrule(lr){6-7}
        Method & \# frames & \makecell{Clench/Unclench} & PnP Once & Full & Cover Cube & Full \\
        \midrule
        $\pi_0$ \citep{black2024pi_0} & 1 & 40.0 & 55.0 & 25.0 & 60.0 & 45.0 \\
        $\pi_0$ \citep{black2024pi_0} & 8 & 40.0 & 60.0 & 55.0 & 65.0 & 45.0 \\
        \textbf{ + \sname~(Ours)} & 8 & \textbf{80.0} & \textbf{75.0} & \textbf{65.0} & \textbf{85.0} & \textbf{60.0} \\
        \midrule
        GR00T N1.5 \citep{bjorck2025gr00t} & 1 & 20.0 & 55.0 & 15.0 & 50.0 & 10.0 \\
        GR00T N1.5 \citep{bjorck2025gr00t} & 8 & \textbf{80.0} & 60.0 & 30.0 & 50.0 & 25.0\\
        \textbf{ + \sname~(Ours)} & 8 & \textbf{80.0} & \textbf{70.0} & \textbf{50.0} & \textbf{55.0} & \textbf{35.0} \\
        \bottomrule
    \end{tabular}\label{tab:main_non_markovian} 
\end{adjustbox}
\end{table*}

\paragraph{Results.}
We find that our framework consistently improves the baselines as shown in \Cref{tab:main_libero,tab:main_simpler,tab:main_robocasa}. This demonstrates that our method indeed can leverage multi-frame visual observations to generate action.
Specifically, we observe that \sname~improves the baselines by a significant margin in Simpler-WidowX Benchmark (\Cref{tab:main_simpler}), a challenging setup due to the Real-to-Sim gap. For instance, \sname~improves the averaged success rates of $\pi_0$ by 14.6\% (41.8\% $\rightarrow$ 56.2\%).
We also observe that our framework improves the performance in the Robocasa Benchmark (\Cref{tab:main_robocasa}). In particular, we observe that our framework improves the performance in Pick and Place (PnP) tasks, consisting of diverse target objects and positions. For instance, \sname~improves the averaged success rates of $\pi_0$-FAST in PnP by 3.0\% (45.5\% $\rightarrow$ 48.5\%).
This demonstrates that the performance gains with \sname~are not specific to the training setup, but instead highlight its ability to generalize across diverse objects and locations.

\begin{figure*}[t]
\vspace{-0.05in}
\centering
\begin{minipage}[t]{0.4\textwidth}
    \centering
    \vspace{0.005in}
    \includegraphics[width=0.89\textwidth]{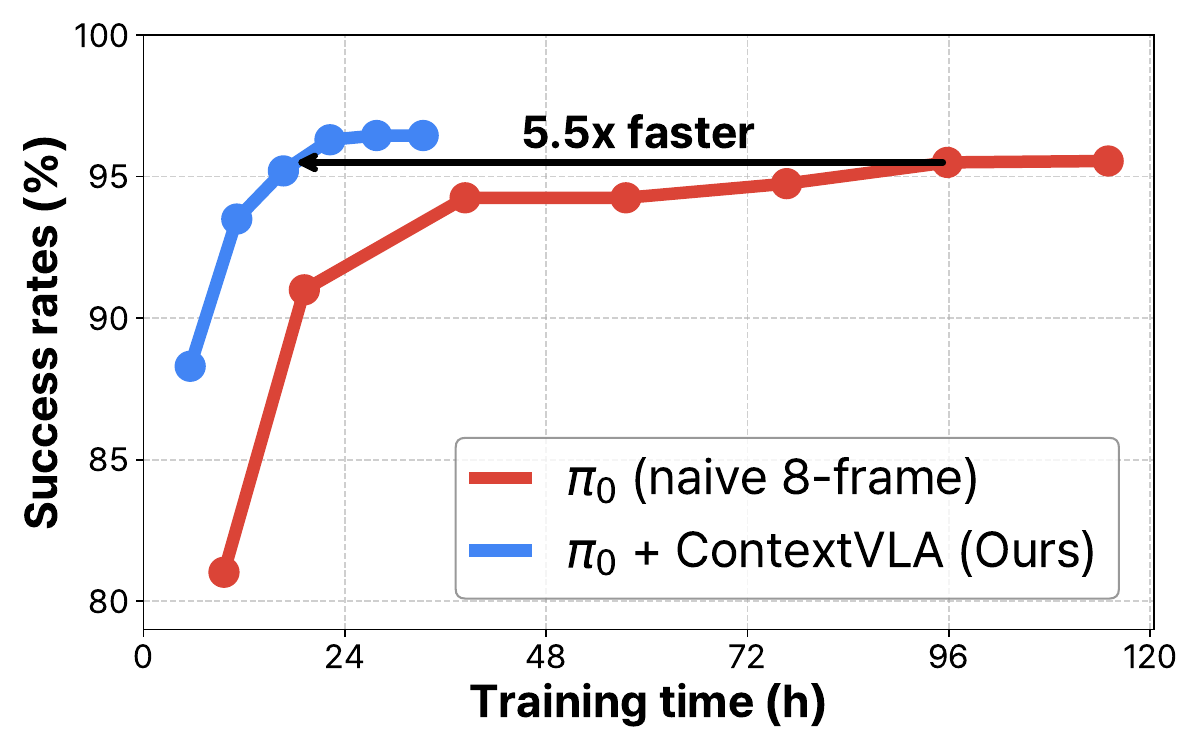}
    \vspace{-0.12in}
    \captionof{figure}{\textbf{Training efficiency.} We report the wall clock time of fine-tuning $\pi_0$ on Libero \citep{liu2023libero} using 4 NVIDIA A100 80GB GPU.}
    \label{fig:wall_time_clock}
    \vspace{-0.06in}
    \captionof{table}{\textbf{Inference efficiency.} We report inference time (ms) required for $\pi_0$ \citep{black2024pi_0} to generate action from 8-frame, 2-view observations using a single NVIDIA A100 80GB GPU. }
    \begin{adjustbox}{max width=0.9\textwidth}
    \begin{tabular}{ccc}
        \toprule
        Compression & KV-caching & Time (ms) \\
        \midrule
        \xk & - & 227.2 \\
        \midrule
        \ck & \xk & 129.9 \\
        \ck & \ck & \phantom{0}\textbf{96.3} \\
        \bottomrule
    \end{tabular}
    \end{adjustbox}
    \label{tab:analysis_efficiency}
\end{minipage} \hfill
\begin{minipage}[t]{0.58\textwidth}
    \centering
    \captionof{table}{\textbf{Component-wise analysis} on Libero \citep{liu2023libero} and Simpler-WidowX \citep{li2024evaluating} benchmarks. All models are $\pi_0$ \citep{black2024pi_0} fine-tuned for 60K iterations with a batch size of 32, using the Libero training dataset for Libero and the BridgeV2 dataset \citep{ebert2021bridge} for Simpler-WidowX. We report the averaged success rates (\%) of the models on each benchmark.}
    \begin{adjustbox}{max width=\textwidth}
    \begin{tabular}{ccc cc}
        \toprule
        \# frames & Depth & Context Token & Libero & Simpler-WidowX \\
        \midrule
        1 & \multicolumn{2}{c}{$\pi_0$ \citep{black2024pi_0}} & 94.6 & 41.8 \\
        \midrule
        8 & \multicolumn{2}{c}{$\pi_0$ \citep{black2024pi_0}} & 95.6 & 47.8 \\
        \midrule
        2 & 2 & \ck & 94.8 & 50.5 \\
        4 & 2 & \ck & 95.0 & 52.0 \\
        8 & 2 & \ck & \textbf{96.5} & \textbf{56.2} \\
        \arrayrulecolor{black!40}\midrule
        8 & 1 & \ck & 96.1 & 46.5 \\
        8 & 2 & \ck & \textbf{96.5} & \textbf{56.2} \\
        8 & 4 & \ck & 96.4 & 53.5 \\
        8 & 6 & \ck & 95.8 & 51.5 \\
        8 & 8 & \ck & 96.4 & 48.5 \\
        \arrayrulecolor{black!40}\midrule
        8 & 2 & \xk & 95.6 & 49.0 \\
        8 & 2 & \ck & \textbf{96.5} & \textbf{56.2} \\
        \arrayrulecolor{black}\bottomrule
    \end{tabular}
    \end{adjustbox}
    \label{tab:component_analysis}
\end{minipage}
\vspace{-0.1in}
\end{figure*}

\subsection{Real-world Robotic Tasks}
To investigate whether our method can leverage multi-frame visual observations, we further evaluate our method on real-world robotic tasks that require temporal context to perform the task successfully (see \Cref{fig:env_nonmark} for examples of tasks used in our experiments).

\paragraph{Tasks.} We design several real-world robotic tasks as follows (see \Cref{appendix:task:real} for more details):
\begin{itemize}[leftmargin=*,itemsep=0mm]
    \item \textbf{Clench/Unclench.} The policy should clench and unclench hand of the humanoid robot repeatedly. At an intermediate state between clenching and unclenching, it must decide to grasp or release the hand depending on the previous movement.
    \item \textbf{PnP Twice.} Given a cube and two plates (A and B), the policy should move the cube from plate A to B and then back from B to A. At each step, it must decide (a) whether to close the gripper (pick) or open it (place), and (b) which plate to place the cube on, based on the previous action.
    \item \textbf{CoverNStack.} Given a cube and two cups, the policy should cover the cube with the closest cup, and then stack the other cup on top of the covered cup. After covering the object with a cup, it must decide which cup to grasp and place it on top of the other cup, depending on the last movement.
\end{itemize}

\paragraph{Training and Evaluation Setup.}
For each task, we collect 50 demonstrations and report the performance by fine-tuning $\pi_0$ \citep{black2024pi_0} and GR00T N1.5 \citep{nvidia2025gr00t} using our framework on the collected demonstrations. We train all models for 30K iterations using the AdamW optimizer with a batch size of 32. We evaluate each model with 20 trials per task and report partial (performs single time for repetitive tasks, and completes only the first subtask for long-horizon tasks) and full success rates. We describe the details of the evaluation in \Cref{appendix:task:real}.

\paragraph{Results.}
In \Cref{tab:main_non_markovian}, we find that \sname~improves the performance of the baselines by a significant margin.
While single-frame baselines often fail to perform simple repetitive actions, \textit{e.g.,} clench/unclench, \sname~consistently outperforms them. This indicates that \sname~leverages temporal context to resolve temporal ambiguities. We observe the same results in sequential tasks (PnP Twice and CoverNStack), \textit{e.g.}, \sname~achieves the highest success rates on both tasks, but single-frame baselines often succeed partially. This further demonstrates the effectiveness of \sname~on long-horizon tasks that require temporal understanding.
In particular, our framework even outperforms VLAs that utilize 8-frame observations without compression.
For instance, in the CoverNStack tasks, fine-tuning $\pi_0$ with \sname~gets a 60\% success rate, whereas fine-tuning it with 8-frame observations without compression gets 45\%. 
The improvement can be attributed to the faster inference speed of our framework, as latency is known to degrade performance during real-world robot deployment \citep{black2025real}.
This highlights the benefit of our approach that compresses the past observations. We provide the qualitative results in \Cref{appendix:qualitative}.

\subsection{Analysis and Ablation Study}

We first perform efficiency analysis in \Cref{fig:wall_time_clock} and \Cref{tab:analysis_efficiency}. After that, in \Cref{tab:component_analysis}, we provide ablation studies to investigate the effect of each component of \sname.

\paragraph{Training Efficiency.}
In \Cref{fig:wall_time_clock}, we analyze how efficient our compression scheme makes VLA when training using multi-frame observations. We compare the success rates under the same training wall-clock time with $\pi_0$ that uses 8-frame visual observations. We find that our framework is much faster to achieve the best performance of the $\pi_0$, \textit{e.g.}, 5.5$\times$ faster on the Libero dataset.

\paragraph{Inference Efficiency.}
In \Cref{tab:analysis_efficiency}, we measure the inference time of our framework to evaluate the inference efficiency of our compression scheme with multi-frame observations. We find that our framework is 2.4$\times$ faster than $\pi_0$ with 8-frame visual observations without compression. In particular, the efficient inference scheme with KV-caching reduces latency by 33.6 ms.

\paragraph{Number of Past Observations.}
We investigate the scalability of \sname~with the number of past observations. We evaluate this by fine-tuning $\pi_0$ \citep{black2024pi_0} with \sname~using different numbers of frames from 2 to 8. We find that the success rates consistently increase as the number of past observations increases. For instance, using 4 frames achieves a success rate of 52.0\% on the Simpler-WidowX benchmark, while using 8 frames achieves a success rate of 56.2\%.

\paragraph{Depth for Token Compression.}
We also investigate the appropriate depth $n$ for compressing past observations. By fine-tuning $\pi_0$ with past observations compressed into a context token at different backbone depths, we find that compression at shallow blocks ($n=2$) is the optimal choice, while compression at other depths still shows the performance improvement. We note that the compression at shallow block allows the model to enhance efficiency during training and inference (see \Cref{tab:analysis_efficiency} and \Cref{fig:wall_time_clock}).

\paragraph{Effect of Global Context Token.} We further investigate whether the compressed context token indeed provides meaningful information to generate actions. To evaluate this, we fine-tune $\pi_0$ by compressing past observations at a middle block and compare two variants: using the compressed tokens in the remaining blocks or discarding them. We find that using context token improves the performance by a large margin, \textit{e.g.}, in Simpler-WidowX, 49.0\% $\rightarrow$ 56.2\%. This demonstrates that the context token captures temporal context from past observations, enabling the policy to generate actions better. We also find that $\pi_0$ using context token even outperforms $\pi_0$ trained on 8-frame observations without compression. We hypothesize this is because the context token summarizes past observations and mitigates redundancy across consecutive frames.

\section{Related work}
\label{sec:related_works}

\paragraph{Vision-Language-Action Models (VLA).}
Recently, VLAs \citep{o2024open,zitkovich2023rt,kim2024openvla} have emerged as a promising framework for general robot policy that can perform many different tasks with a single model, by leveraging pre-trained Vision-Language Model (VLM) \citep{liu2023visual,beyer2024paligemma,bai2025qwen2,chen2025eagle} and training on large-scale robot manipulation datasets \citep{ebert2021bridge,walke2023bridgedata,o2024open,khazatsky2024droid,bu2025agibot}. However, many existing VLAs are trained to process only a single-frame visual observation to generate actions \citep{kim2024openvla,li2024cogact,black2024pi_0,shukor2025smolvla,yang2025magma,bjorck2025gr00t,hung2025nora,qu2025spatialvla,driess2025knowledge,nvidia2025gr00t,cheang2025gr}, limiting their ability to perform diverse robotic tasks that require temporal context. To address this, several recent approaches have attempted to leverage multi-frame observations. A common strategy is to use full context of multi-frame observations to generate actions \citep{wu2023unleashing,team2024octo,cheang2024gr,huang2025otter,liu2025towards,wang2025unified}. However, using multi-frame observations without compression increases computation and memory overheads by increasing input dimensionality that VLA processes, resulting in inefficient training and inference. In contrast, a recent approach, TraceVLA \citep{zheng2024tracevla}, summarizes the observations by tracking the robot trace. However, it requires external point tracking model \citep{karaev2024cotracker}, making inference still slow. In this paper, we introduce an efficient VLA framework that leverages multi-frame visual observations.

\paragraph{Efficient Training and Inference of Multi-frame Policy.}
A higher input dimensionality compared to single-frame observations makes the training and inference of policy model inefficient. This hinders scaling up the policy model training \citep{black2024pi_0,bjorck2025gr00t}, and poses a significant challenge for deployment, as inference speed is a critical issue in robotics \citep{black2025real}. To address this, some approaches handle multi-frame inputs efficiently by compressing past observations \citep{wen2020fighting,seo2023regularized}, selecting a key-frames \citep{wen2021keyframe}, or summarizing entire sequences into high-level visualizations \citep{sundaresan2024rt,zheng2024tracevla}. In addition, recent work proposes two-stage scheme where it first trains a single-frame policy, and then trains a multi-frame policy after freezing visual encoder \citep{torne2025learning}. Our work also proposes a method that allows the policy to leverage multi-frame observation more efficiently that compressing past observations into a single token.
\section{Conclusion}
\label{sec:conclusion}
In this work, we have presented \sname, an efficient framework for Vision-Language-Action models (VLA) that leverages multi-frame visual observations for action generation. Motivated by the observation that VLAs mitigate the performance degradation suffered by behavior cloning policies with multi-frame inputs, we introduce a simple yet effective method that compresses past observations into a single context token, allowing the VLA to capture temporal context more efficiently. Our experiments show that \sname~leverages multi-frame observations to improve the performance of existing VLAs. We also find that our scheme retains the benefits of multi-frame training with less training and inference time. We hope that our work facilitates future research toward generalist robot policies that can capture temporal context to perform more diverse tasks.

\section*{Reproducibility Statement}
For the reproducibility of our results, we provide the implementation details in \Cref{appendix:implementation,appendix:task}, including training and inference setups. In addition, we will open-source the source code with the model checkpoint.

\section*{Acknowledgments}
This paper was supported by RLWRLD. We thank Jimin Lee and Myungkyu Koo for providing
helpful support in conducting real-world experiments.

\bibliography{iclr2026_conference}
\bibliographystyle{iclr2026_conference}

\newpage

\appendix

\section{Implementation Details}\label{appendix:implementation}
We report the performance of \sname, by fine-tuning pre-trained Vision-Language-Action models (VLA). Specifically, we use $\pi_0$ \citep{black2024pi_0}, GR00T N1.5 \citep{bjorck2025gr00t}, and $\pi_0$-FAST \citep{pertsch2025fast} by following their official implementation.
We use 8 consecutive frames as multi-frame observations, and we compress past observations into context tokens  at the output of the 2nd VLM block (\textit{i.e.}, $n$=2). For simulated tasks, we train $\pi_0$ + \sname, $\pi_0$-FAST + \sname, and GR00T N1.5 + \sname~for 60K, 30K, and 60K iterations, respectively. For real-world tasks, we only need to train the model on a single task with 50 demonstrations, so we train $\pi_0$ + \sname~and GR00T N1.5 + \sname~for fewer (30K) iterations.

\vspace{-0.1in}

\begin{table}[h]
\centering
\caption{Hyperparameter details of training \sname~in simulated robotic tasks}
\vspace{-0.1in}
\begin{adjustbox}{max width=\textwidth}
\begin{tabular}{lccc}
\toprule
 & $\pi_0$ + \sname & $\pi_0$-FAST + \sname & GR00T N1.5 + \sname \\
\midrule
optimizer & AdamW & AdamW & AdamW\\
optimizer momentum & $\beta_1$, $\beta_2$ = 0.9, 0.95 & $\beta_1$, $\beta_2$ = 0.9, 0.95 & $\beta_1$, $\beta_2$ = 0.95, 0.999 \\
optimizer weight decay & 1e-10 & 1e-10 & 1e-5 \\
learning rate & 2.5e-5 & 2.5e-5 & 1e-4\\
learning rate scheduler & Cosine decay & Cosine decay & Cosine decay \\
warmup iterations & 1000 & 1000 & 3000 \\
batch size & 32 & 32 & 32 \\
training iterations (simulated tasks) & 60000 & 30000 & 60000 \\
training iterations (real-world tasks)  & 30000 & - & 30000 \\
\bottomrule
\end{tabular}
\end{adjustbox}
\end{table}

\section{Benchmark Details}\label{appendix:task}

\subsection{Simulated Robotic Manipulation Benchmarks} \label{appendix:task:simul}
We evaluate our method on the following simulated robotic manipulation benchmarks.

\paragraph{Libero.} We consider Libero benchmark \citep{liu2023libero}, a widely used simulated robotic manipulation benchmark for evaluating the performance of Vision-Language-Action models (VLA). It uses a Franka robot. It consists of 4 different sub-benchmarks (Spatial, Object, Goal, and Long), each comprising 10 tasks. While each sub-benchmark has its own training dataset \citep{team2024octo,kim2024openvla}, we follow the setup of $\pi_0$ \citep{black2024pi_0} that combines all training datasets to train the models and reports the performance separately for each sub-benchmark. We use a fixed front-view camera and a wrist camera of 224$\times$224 resolution without depth. We use the end-effector position as the action mode. We evaluate models 50 times for each task by varying the position of objects, and report the average success rates.

\paragraph{Simpler-WidowX.} We consider Simpler-WidowX Benchmark \citep{li2024evaluating}, a more challenging simulated benchmark. It uses the WidowX robot. It consists of 4 tasks (Spoon on Towel, Carrot on Plate, Stack Cube, and Put Eggplant in Basket). We follow the common setup \citet{li2024evaluating} that uses the BridgeV2 dataset \citep{walke2023bridgedata} to fine-tune the model, and then evaluate the models in this benchmark. We use the primary camera in the BridgeV2 dataset when training, and use a fixed third-person view camera of 224$\times$224 resolution without depth. We use the end-effector position as the action mode. For each task, we evaluate models 150 times for each task by varying the random seed of this benchmark, and report the average success rates.

\paragraph{Robocasa.} We additionally consider Robocasa benchmark \citep{nasiriany2024robocasa}, which includes 24 tasks in simulated kitchen environments. We randomly sample 100 demonstrations per task in the machine-generated training dataset in Robocasa and combine all these demonstrations to train the models. We use a fixed left-view camera, a fixed right-view camera, and a wrist camera of 224$\times$224 resolution without depth. We use the end-effector position as the action mode. We evaluate models 50 times for each task with random seeds, and report the average success rates.

\newpage

\subsection{Real-world Robotic Tasks} \label{appendix:task:real}
We design several real-world robotic tasks. We collect 50 demonstrations per task and report the performance of the models by fine-tuning VLAs on each task. We evaluate models 20 times for each task. In the below, we describe the detailed setups (see \Cref{fig:appendix_real_demo} for a visualization of the collected train demonstrations):

\begin{figure}[h]
\vspace{-0.15in}
\centering
\begin{subfigure}[t]{\textwidth}
  \centering
  \includegraphics[width=.9\linewidth]{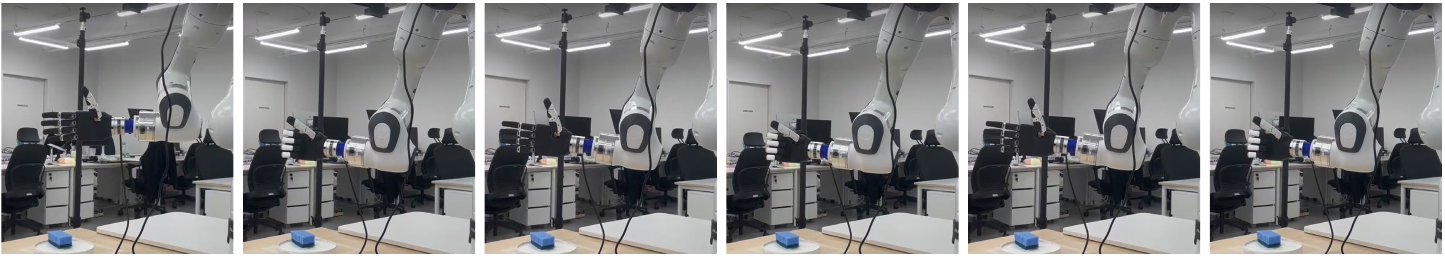}
\vspace{-0.07in}
  \subcaption{Clenching and unclenching hand (Clench/Unclench)}
\end{subfigure}
\begin{subfigure}[t]{\textwidth}
  \centering
  \includegraphics[width=.9\linewidth]{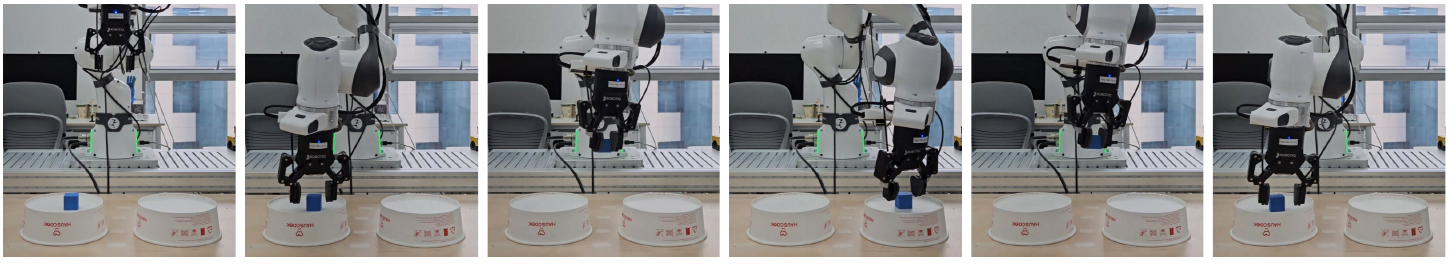}
\vspace{-0.07in}
  \subcaption{Pick and place twice (PnP Twice)}
\end{subfigure}
\begin{subfigure}[t]{\textwidth}
  \centering
  \includegraphics[width=0.9\linewidth]{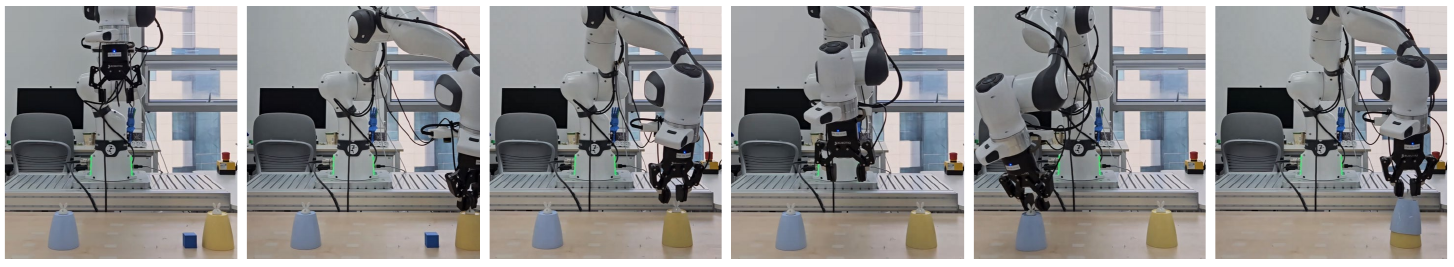}
\vspace{-0.07in}
  \subcaption{Cover the cube and stack up (CoverNStack)}
\end{subfigure}
\vspace{-0.1in}
\caption{\textbf{Visualization of real-world robotic task demonstrations.}
}
\label{fig:appendix_real_demo}
\end{figure}

\paragraph{Clench/Unclench.} The policy should clench and unclench the right hand of a humanoid robot repeatedly. For the robot setup, we use Franka Research 3 (Robot arm) + Inspire Dex hands RH56FTP (Robot hand). We use the third-person view camera and wrist camera of 224$\times$224 resolution without depth. We use an absolute end-effector position as the action mode. We use the text instruction ``\textit{Clench and then unclench the hand repeatedly.}" for this task. We report success if the robot clenching and unclenching a hand at least five times within 1 minute.

\paragraph{PnP Twice.} Given a cube and two plates (A and B), the policy should move the cube from plate A to B and then back from B to A. We use Franka Research 3 (Robot arm) + DROID gripper. We use the third-person view camera and wrist camera of 720$\times$1280 resolution without depth. We use the visual observation after resizing it to 224$\times$224 resolution. We use absolute joint position as action mode. We use the text instruction ``\textit{Pick up the cube and place it on the opposite side, and then return it to the original side.}" for this task. We report partial success if the robot move the cube to the opposite side once, and full success if the robot complete the task within 1 minute.

\paragraph{CoverNStack.} Given a cube and two cups, the policy should cover the cube with the closest cup, and then stack the other cup on top of the covered cup. We use the same robot, camera setup, and action mode as in the PnP Twice task. We use the text instruction ``\textit{Cover the cube with the nearest cup, then stack the other cup on top of it.}" for this task. We report partial success if the robot covers the cube, and full success if the robot complete the task within 1 minute.

\newpage

\section{Baselines} \label{appendix:baselines}
We describe the main idea of baseline methods that we used for the evaluation.
\begin{itemize}[leftmargin=*,itemsep=0mm]
    \item \textbf{Octo} \citep{team2024octo} processes visual observations and task instructions to produce a read-out token, which is fed into a diffusion model to generate actions.
    \item \textbf{OpenVLA} \citep{kim2024openvla} fine-tunes a pretrained VLM to autoregressively generate action tokens from visual observations and task instructions.
    \item \textbf{RoboVLMs} \citep{liu2025towards} uses multi-frame observations to generate action. It obtains tokens from each frame via the VLM and then concatenates them to generate the action.
    \item \textbf{TraceVLA} \citep{zheng2024tracevla} first extracts an image of the robot trajectories from multi-frame observation using point tracking models, and then uses this image to generate action via autoregressive modeling.
    \item \textbf{SpatialVLA}  \citep{qu2025spatialvla} integrates 3D spatial understanding capability into VLAs by using Ego3D position encoding and adaptive action grids.
    \item \textbf{NORA} \citep{hung2025nora} uses Qwen2.5-VL \citep{bai2025qwen2} as a backbone model to generate discrete action tokens. It decodes the token to action values using the FAST tokenizer.
    \item \textbf{$\pi_0$} \citep{black2024pi_0} proposes a diffusion-based VLA that shares self-attention layers between VLM and diffusion transformer.
    \item \textbf{$\pi_0$-FAST} \citep{pertsch2025fast} is an autoregressive VLA utilizing Frequency-space Action Sequence Tokenization (FAST) action tokenizer. The tokenizer applies the discrete cosine transform (DCT) algorithm for encoding the continuous action values to discrete tokens. Then, the FAST tokenizer applies Byte Pair Encoding (BPE) to compress sequences.
    \item \textbf{GR00T N1.5} \citep{nvidia2025gr00t} also proposes diffusion-based VLA, but it feeds the last hidden states of the VLM into the cross-attention layer of the diffusion transformer.
\end{itemize}

\section{Detailed Action Generation Process} \label{appendix:action_generation}

We here describe the detailed process of action generation of VLAs used in our framework: Autoregressive modeling \citep{kim2024openvla,pertsch2025fast}, and diffusion-based modeling \citep{black2024pi_0,bjorck2025gr00t}.

\subsection{Autoregressive Modeling.}
Autoregressive VLAs \citep{o2024open, pertsch2025fast} encode a continuous robot action $\rva_{t:t+l}$ into discrete action tokens using an action tokenizer \citep{bu2025univla,pertsch2025fast}, and then train a Vision-Language Model (VLM) to generate the discrete action tokens via next token prediction. Concretely, We first use the visual observations $\rvo_{t-k:t}$ and the text instruction $\rvc_t$ as a prompt for the VLM. And then, VLM autoregressively generates tokens, where we consider the tokens as discretized action values. After the tokens are generated, we decode the tokens into continuous action by using action de-tokenizer.

In our experiments, we use $\pi_0$-FAST, which uses FAST tokenizer \citep{pertsch2025fast} for encoding and decoding actions. The FAST tokenizer improves the compactness and expressivity of the discretized action space by tokenizing actions in the frequency domain. Specifically, actions are first transformed by a discrete cosine transform (DCT; \citealt{ahmed2006discrete}), then quantized and compressed into discrete tokens via byte-pair encoding (BPE; \citealt{gage1994new}). These tokens are assigned to unused special tokens in the vocabulary of the VLM for training and generation.

\newpage

\subsection{Diffusion-based Modeling.}
Diffusion-based VLAs \citep{black2024pi_0,bjorck2025gr00t} generates action using diffusion model conditioned on a VLM. We first process the visual observation $\rvo_{t-k:t}$ and text instruction $\rvc_t$ to extract high-level semantic information using VLM. Then, the extracted features are used as conditioning for Diffusion Transformer (DiT; \citealt{peebles2022scalable}). Here, there are many approaches to condition the features on DiT, \textit{e.g.}, $\pi_0$ \citep{black2024pi_0} shares self-attention layers between VLM and DiT, and GR00T N1.5 \citep{bjorck2025gr00t} feeds the last hidden states of the VLM into cross-attention layer of DiT.

During Training, we sample denoising timestep $\tau \in [0,1]$. Then, for the target action chunk $\rva_{t:t+l}$ of the robot \citep{zhao2023learning, chi2023diffusion}, we add noise to the action using Gaussian noise $\epsilon \sim \gN (\mathbf{0}, \mathbf{I})$:
\begin{align}
 \rva_{t:t+l}^\tau = \tau \rva_{t:t+l} + (1-\tau)\epsilon.
\end{align}
Then, VLA model $\pi_\theta$ processes visual observation $\rvo_t$ and text instruction $\rvc_t$ to approximate the denoising direction, \textit{i.e.}, $\epsilon- \rva_{t:t+l}$ by minimizing flow-matching loss:
\begin{align}
 \mathcal{L} = \mathbb{E}_{\tau}{\bigg[\big\|\pi_\theta(\rva_{t:t+l}^\tau|\rvo_{t-k:t},\rvc_t)-(\epsilon - \rva_{t:t+l})\big\|^2\bigg]}.
\end{align}

During inference, we generate action $\rva_{t:t+l}$ through denoising $N$ steps. We sample random noise $\rva_{t:t+l}^0 \sim \mathcal{N}(\mathbf{0}, \mathbf{I})$, and apply ODE or SDE sampler to denoise it to generate action $\rva_{t:t+l}$.

\section{Weight-Initialization Analysis}
\begin{figure}[h]
\centering
\begin{subfigure}[h]{0.48\textwidth}
  \centering
  \includegraphics[width=0.95\linewidth]{assets/init_robomimic_square.pdf}
  \vspace{-0.07in}
  \subcaption{Square task from robomimic benchmark}
  \label{fig:appendix:init_robomimic}
\end{subfigure}
\begin{subfigure}[h]{0.48\textwidth}
  \centering
  \includegraphics[width=0.95\linewidth]{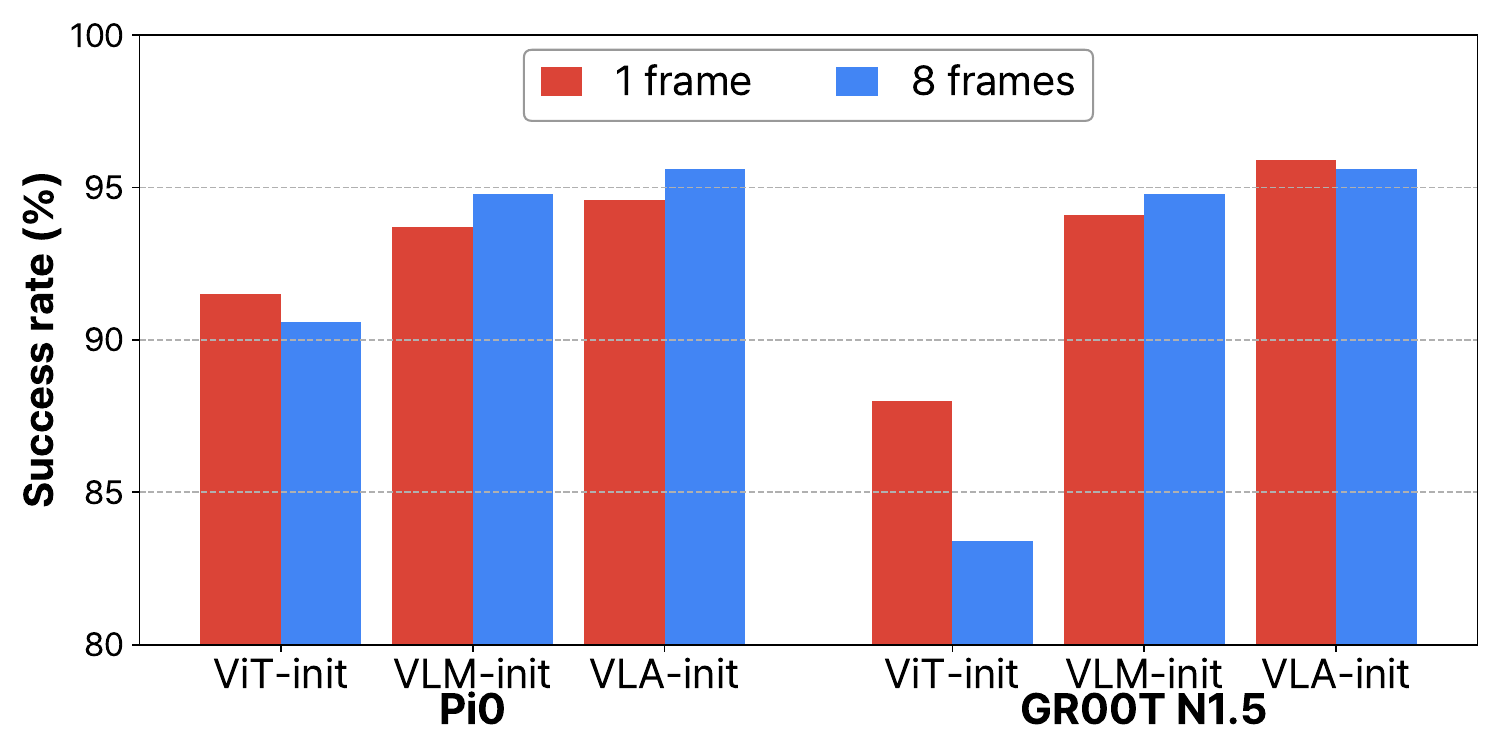}
  \vspace{-0.07in}
  \subcaption{Libero benchmark}
  \label{fig:appendix:init_libero}
\end{subfigure}
\caption{Success rates (\%) of policy models of VLA architecture trained using either 1-frame or 8-frame observations under different weight initialization schemes. ViT, VLM, and VLA-init indicate how the VLA architecture is initialized for training; we use a pre-trained vision encoder, VLM, or VLA, respectively, and other parameters are randomly initialized.
}
\label{fig:appendix:motivation}
\end{figure}

In \Cref{fig:appendix:motivation}, we analyze how different parameter initializations affect policy performance when using multi-frame observation. We consider two architectures, $\pi_0$ \citep{black2024pi_0} and GR00T N1.5 \citep{nvidia2025gr00t}, and train them under three initialization schemes:
(1) For ViT-init, only a vision encoder is initialized with a pre-trained model \citep{zhai2023siglip,tschannen2025siglip2}. (2) For VLM-init, the vision encoder and VLM backbone are initialized with a pre-trained VLM \citep{beyer2024paligemma,chen2025eagle}. (3) For VLA-init, entire model is initialized with a pre-trained VLA, \textit{i.e.}, $\pi_0$ or GR00T N1.5.
We observe that a policy model of VLA architecture but initialized only with a vision encoder (\textit{i.e.}, ViT-init) suffers from performance degradation when using multi-frame inputs. In contrast, policies initialized with a pre-trained VLM or with VLA alleviate or even overcome this issue, indicating the key factor in mitigating the problem is leveraging pre-trained Vision-Language Model (VLM) to extract information for action generation.

\newpage

\section{Qualitative Results} \label{appendix:qualitative}
\begin{figure}[h]
\centering
\includegraphics[width=1.0\textwidth]{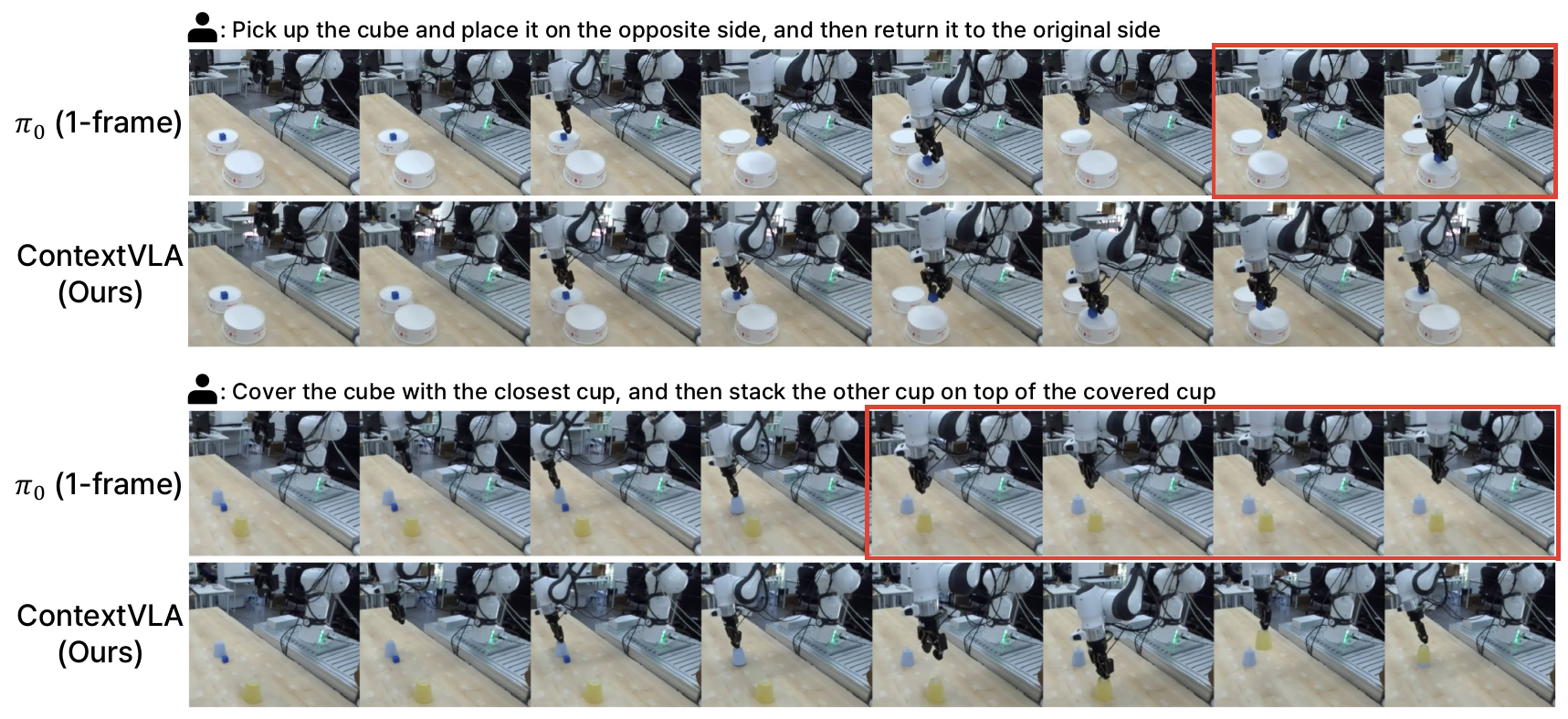}
\vspace{-0.2in}
\caption{\textbf{Qualitative results.} $\pi_0$ that uses single-frame observations fails to determine the correct next action due to the lack of utilizing temporal context (in red box), but \sname~leverages temporal context to determine the correct next action depending on the previous movement of a robot.}
\label{fig:appendix_qualitative}
\end{figure}

In \Cref{fig:appendix_qualitative}, we provide qualitative results for the real-world robotic tasks. We find that a policy that uses single-frame observations often fails to determine the correct next action, leading to failure in many cases, but \sname~leverages multi-frame observations to perform the task successfully.

\section{Additional Quantitative Results} \label{appendix:quantitative}

In \Cref{tab:appendix_libero,tab:appendix_simpler}, we report the performance of our method compared to the VLAs that use 8-frame as visual observations. We find that \sname~achieves performance comparable to, or even surpassing, the VLAs that use 8-frame observation without compression. This indicates that \sname~effectively compresses past observations into an amortized token that captures the temporal context well.

\section{Use of AI Tools} 
We acknowledge that a large language model (LLM) was used to refine the phrasing and grammar of the manuscript.

\newpage

\begin{table*}[h]
\caption{\textbf{Results on Libero.} We report the success rates (\%) of various VLAs fine-tuned on the training dataset of Libero \citep{liu2023libero}.}
\centering\small
\vspace{-0.1in}
\begin{adjustbox}{max width=0.95\textwidth}
    \begin{tabular}{lc ccccc}
        \toprule
        Method & \# frames & Spatial & Object & Goal & Long & Avg. \\
        \midrule
        $\pi_0$ \citep{black2024pi_0} & 1 & 96.0 & 97.2 & 96.0 & 89.2 & 94.6 \\
        $\pi_0$ \citep{black2024pi_0} & 8 & 97.2 & \textbf{98.4} & 94.6 & 92.0 & 95.6 \\
        \textbf{ + \sname~(Ours)} & 8 & \textbf{97.4} & 98.2 & \textbf{96.4} & \textbf{93.8} & \textbf{96.5} \\
        \midrule
        $\pi_0$-FAST \citep{pertsch2025fast} & 1 & 96.6 & 96.6 & 95.2 & 85.2 & 93.4 \\
        $\pi_0$-FAST \citep{pertsch2025fast} & 8 & 98.0 & \textbf{99.2} & \textbf{96.8} & \textbf{91.2} & \textbf{96.3} \\
        \textbf{ + \sname~(Ours)} & 8 & \textbf{98.2} & \textbf{99.2} & 95.6 & 90.2 & 95.8 \\
        \midrule
        GR00T N1.5 \citep{nvidia2025gr00t} & 1  & 98.3 & \textbf{99.4} & 96.7 & 89.0 & 95.9 \\
        GR00T N1.5 \citep{nvidia2025gr00t} & 8  & \textbf{98.4} & \textbf{99.4} & 95.8 & 88.6 & 95.6 \\
        \textbf{ + \sname~(Ours)} & 8  & \textbf{98.4} & 99.0 & \textbf{97.2} & \textbf{93.4} & \textbf{97.0} \\
        \bottomrule
    \end{tabular}\label{tab:appendix_libero} 
\end{adjustbox}
\end{table*}

\begin{table*}[h]
\caption{\textbf{Results on Simpler-WidowX.} We report the success rates (\%) of various VLAs fine-tuned on the Bridgev2 dataset \citep{walke2023bridgedata}.}
\centering\small
\vspace{-0.1in}
\begin{adjustbox}{max width=0.95\textwidth}
    \begin{tabular}{lc ccccc}
        \toprule
        Method & \# frames & \shortstack{Spoon\\on Towel} & \shortstack{Carrot\\on Plate} & \shortstack{Stack\\Cube} & \shortstack{Put Eggplant\\ in Basket} & Avg. \\
        \midrule
        $\pi_0$ \citep{black2024pi_0} & 1 & 46.5 & 38.7 & 42.7 & 39.3 & 41.8 \\
        $\pi_0$ \citep{black2024pi_0} & 8 & 41.3 & 42.7 & \textbf{43.3} & 64.0 & 47.8 \\
        \textbf{ + \sname~(Ours)} & 8 & \textbf{53.3} & \textbf{56.0} & 41.3 & \textbf{74.0} & \textbf{56.2} \\
        \midrule
        $\pi_0$-FAST \citep{pertsch2025fast} & 1 & 59.0 & 79.0 & 65.0 & 33.0 & 59.0 \\
        $\pi_0$-FAST \citep{pertsch2025fast} & 8 & 58.0 & 58.0 & \textbf{90.0} & 52.0 & 64.5 \\
        \textbf{ + \sname~(Ours)} & 8 & \textbf{60.7} & \textbf{81.3} & 78.7 & \textbf{62.0} & \textbf{70.7} \\
        \midrule
        GR00T N1.5 \citep{nvidia2025gr00t} & 1 & \textbf{30.0} & 28.0 & \textbf{16.0} & \phantom{0}42.7 & 29.2 \\
        GR00T N1.5 \citep{nvidia2025gr00t} & 8 & \phantom{0}8.0 & \phantom{0}2.0 & \phantom{0}2.0 & \phantom{0}\phantom{0}8.0 & \phantom{0}5.0\\
        \textbf{ + \sname~(Ours)} & 8 & 28.0 & \textbf{29.3} & 14.7 & \phantom{0}\textbf{50.3} & \textbf{31.8} \\
        \bottomrule
    \end{tabular}\label{tab:appendix_simpler} 
\end{adjustbox}
\end{table*}

\end{document}